\documentclass[11pt,a4paper]{article}
\usepackage[hyperref]{acl2019}
\usepackage{times}
\usepackage{latexsym}

\usepackage{placeins}
\usepackage{booktabs}
\usepackage{appendix}

\usepackage{svg}
\usepackage{amsmath}

\usepackage{bbm}
\usepackage{graphicx}  
\usepackage[utf8]{inputenc}
\usepackage{booktabs}
\usepackage{url}
\usepackage{multirow}
\aclfinalcopy 

\title{An Empirical Study on Model-agnostic Debiasing Strategies \\
for Robust Natural Language Inference}

\author{Tianyu Liu$^1$ \Thanks{ Equal contribution.} \ \ \ \ Xin Zheng$^3$ \footnotemark[1] \ \ \ Xiaoan Ding$^4$ \ \ \ Baobao Chang$^1$ $^2$ \ \ \ Zhifang Sui$^1$ $^2$ \\
 $^1$ Peking University, Beijing, China \ \  $^2$ Peng Cheng Laboratory, Shenzhen, China \\
 $^3$ Beijing University of Posts and Telecommunications, Beijing, China \\
 $^4$ University of Chicago, IL, USA \\
 \texttt{\{tianyu0421, chbb, szf\}@pku.edu.cn,} \
  \texttt{zheng\_xin@bupt.edu.cn} \\
   \texttt{xiaoanding@uchicago.edu}}

\date{}

\begin{document}
\maketitle
\begin{abstract}
The prior work on natural language inference (NLI) debiasing mainly targets at one or few known biases while not necessarily making the models more robust.
In this paper, we focus on the model-agnostic debiasing strategies and explore how to (or is it possible to) make the NLI models robust to multiple distinct adversarial attacks while keeping or even strengthening the models' generalization power.
We firstly benchmark prevailing neural NLI models including pretrained ones on various adversarial datasets.
We then try to combat distinct known biases by modifying a mixture of experts (MoE) ensemble method \cite{clark2019don} and show that it's nontrivial to mitigate multiple NLI biases at the same time, and that model-level ensemble method outperforms MoE ensemble method.
We also perform data augmentation including text swap, word substitution and paraphrase and prove its efficiency in combating various (though not all) adversarial attacks at the same time.
Finally, we investigate several methods to merge heterogeneous training data (1.35M) and perform model ensembling, which are straightforward but effective to strengthen NLI models.
\end{abstract}

\section{Introduction}
Natural language inference (NLI) (also known as recognizing textual entailment) is a widely studied task which aims to infer the relationship 
(e.g., \emph{entailment}, \emph{contradiction}, \emph{neutral})
between two fragments of text, known as \emph{premise} and \emph{hypothesis} \cite{dagan2006pascal,DBLP:series/synthesis/2013Dagan}. 
Recent works have found that NLI models are sensitive to the compositional features \cite{nie2018analyzing}, syntactic heuristics \cite{mccoy2019right}, stress test \cite{geiger2018stress,DBLP:conf/coling/NaikRSRN18} and human artifacts in the data collection phase \cite{gururangan2018annotation,poliak2018hypothesis,tsuchiya2018performance}.
Accordingly, several adversarial datasets are proposed for these known biases\footnote{In this paper, we use the term `bias' to refer to these known dataset biases in NLI following \citet{clark2019don}. In other context, `bias' may refer to systematic mishandling of gender or evidences of racial stereotypes \cite{RudingerMD17} in NLI datasets or models.}.  

Through our preliminary trials on specific adversarial datasets, we find that although the model specific or dataset specific debiasing methods could increase the model performance on the paired adversarial dataset, they might hinder the model performance on other adversarial datasets, as well as hurt the model generalization power, i.e. deficient scores on cross-datasets or cross-domain settings. 
These phenomena motivate us to investigate if it exists a unified model-agnostic debiasing strategy which can mitigate distinct (or even all) known biases while keeping or strengthening the model generalization power.

We begin with NLI debiasing models.
To make our trials more generic, we adopt a mixture of experts (MoE) strategy \cite{clark2019don}, which is known for being model-agnostic and is adaptable to various kinds of known biases, as backbone. Specifically we treat three known biases, namely word overlap, length mismatch and partial input heuristics as independent experts and train corresponding debiasing models. 
Our results show that the debiasing methods tied to one particular known bias may not be sufficient to build a generalized, robust model. This motivates us to investigate a better solution to integrate the advantages of distinct debiasing models. We find model-level ensemble is more effective than other MoE ensemble methods. Although our findings are based on the MoE backbone due to the prohibitive exhaustive studies on the all existing debiasing strategies, we provide actionable insights on combining distinct NLI debiasing methods to the practitioners.

Then we explore model agnostic and generic data augmentation methods in NLI, including text swap, word substitution and paraphrase. We find these methods could help NLI models combat multiple (though not all) adversarial attacks, e.g. augmenting training data by swapping hypothesis and premise could boost the model performance on stress tests and lexical inference test, and data augmentation by paraphrasing the hypothesis sentences could help the models resist the superficial patterns from syntactic and partial input heuristics.
We also observe that increasing training size by incorporating heterogeneous training resources is a simple but effective method to build robust and generalized models. Specifically we investigate how to incorporate  different training data with different sizes and annotation processes, as well as the best way to perform model ensembling.

\begin{table}[]
\centering
\small
\begin{tabular}{c|c|c|c|c}
\hline
Datasets & Paper & Categories & Labels & Size \\\hline
PI-CD & (a) & $1\|3\|7$ & (E,N,C) & 3.2k\\
PI-SP & (b) & $1\|3\|7$ & (E,N,C) & .37k\\
IS-SD & (c) & $2\|5\|8$ & ($\neg$E, E) & 30k\\
IS-CS & (d) & $2\|3\|7$ & (E,N,C) & .65k\\
LI-LI & (e)(f) & $2\|4\|9$ & (E,C) & 9.9K\\
LI-TS & (g)(h) & $2\|6\|10$ &  ($\neg$C, C) & 9.8K\\
ST-WO & (e) & $2\|4\|11$ & (E,N,C) & 9.8K\\
ST-NE & (e) & $2\|4\|11$ &  (E,N,C) & 9.8K\\
ST-LM & (e) & $2\|4\|11$ &  (E,N,C) & 9.8K\\
ST-SE & (e) & $2\|4\|12$ &  (E,N,C) & 31K\\ \hline
\end{tabular}
\\
\begin{tabular}{ll}
(a) \citet{gururangan2018annotation} & (b) \citet{liu2020hyponli} \\
(c) \citet{mccoy2019right}     & (d) \citet{nie2018analyzing}\\
(e) \citet{DBLP:conf/coling/NaikRSRN18} & (f) \citet{GlocknerSG18}\\
(g) \citet{wang2019if} & (h) Minervini and Riedel
\end{tabular}

\begin{tabular}{l|c|c}
\hline
Category & First-level & Second-level  \\\hline
1 & $(\mathbf{\uppercase\expandafter{\romannumeral1}})$ & Partial input heuristics \\
2 & $(\mathbf{\uppercase\expandafter{\romannumeral1}})$ & Inter-sentence heuristics \\ \hline
3 & $(\mathbf{\uppercase\expandafter{\romannumeral2}})$ & Instance selection \\
4 & $(\mathbf{\uppercase\expandafter{\romannumeral2}})$ & Single Sentence Modification \\
5 & $(\mathbf{\uppercase\expandafter{\romannumeral2}})$ & Sentence Pair Modification \\
6 & $(\mathbf{\uppercase\expandafter{\romannumeral2}})$ & Sentence Pair Swapping \\ \hline
7 & $(\mathbf{\uppercase\expandafter{\romannumeral3}})$ & Lexical Statistical Irregularity \\
8 & $(\mathbf{\uppercase\expandafter{\romannumeral3}})$ & Syntactic Statistical Irregularity \\
9 & $(\mathbf{\uppercase\expandafter{\romannumeral3}})$ & Lexical Inference \\
10 & $(\mathbf{\uppercase\expandafter{\romannumeral3}})$ & First Order Logic \\
11 & $(\mathbf{\uppercase\expandafter{\romannumeral3}})$ & Stress Test - Distraction Test \\
12 & $(\mathbf{\uppercase\expandafter{\romannumeral3}})$ & Stress Test - Noise Test \\ 
\hline 
\end{tabular}
\\ $(\mathbf{\uppercase\expandafter{\romannumeral1}})$ Where are the heuristics? \\
$(\mathbf{\uppercase\expandafter{\romannumeral2}})$ How did the dataset constructed?\\
$(\mathbf{\uppercase\expandafter{\romannumeral3}})$ Which aspect did the dataset detect? \\
\vspace{-0.2cm}
\caption{The information of adversarial datasets (Sec \ref{sec:bencmark}) we use in this paper. We categorize and rename these datasets as discussed in Sec \ref{tab:adversarial_datasets}.
}
\label{tab:data_categorize}
\vspace{-0.3cm}
\end{table}

\section{Benchmark Datasets} \label{sec:bencmark}
Our benchmark datasets include the adversarial datasets\footnote{Some datasets listed in Table \ref{tab:data_categorize} were originally proposed to probe for systematicity. Here we call them `adversarial' datasets in the sense that the NLI models can not reach the same performance on these datasets as the in-domain test sets.} and some widely used general-purpose NLI datasets which test the generalization power of NLI models. \footnote{The datasets used in this paper can be found in the following github repository \url{https://github.com/tyliupku/nli-debiasing-datasets}}

\subsection{Adversarial Datasets}\label{tab:adversarial_datasets}

\noindent \textbf{Categorization}: 
to provide more insights on how the adversarial datasets attack the models, we roughly categorize them in Table \ref{tab:data_categorize} according to their characteristics and elaborate the categorization in this section. To facilitate the narrative of following sections, we rename the adversarial datasets according to their prominent features. 

\noindent \textbf{Comparability}: all the following datasets are collected based on the public available resources proposed by their authors, thus the experimental results in this paper are comparable to the numbers reported in the original papers and the other papers that use these datasets\footnote{The ownership of these datasets belong to their authors. We encourage the readers to acknowledge and cite the original papers listed in Table \ref{tab:data_categorize} when using them.}.

\subsubsection{Partial-input (PI) Heuristics}\label{sec:partial-input}
Partial-input heuristics refer to the hypothesis-only bias \cite{poliak2018hypothesis} in NLI.

\noindent \textbf{Classifier Detected Datasets (PI-CD)}: \citet{gururangan2018annotation} trained a neural classifier (fastText\footnote{\url{https://fasttext.cc/}}) on the hypothesis sentences and then treated those instances in the SNLI test sets which can not be correctly classified as `hard' instances.

\noindent \textbf{Surface Pattern Datasets (PI-SP)}: \citet{liu2020hyponli} recognized surface patterns which are highly correlated to the specific labels and correspondingly proposed adversarial test sets which are against surface patterns' indications. We use their `hard' instances for MultiNLI mismatched dev set as adversarial datasets. 
\subsubsection{Inter-sentences (IS) Heuristics}\label{sec:inter-sentence}

\noindent \textbf{Syntactic Diagnostic Datasets (IS-SD)}: The HANS dataset \cite{mccoy2019right} includes lexical overlap, subsequence and constituent heuristics between the hypothesis and premises sentences, e.g. the model might incorrectly predict `\emph{entailment}' for instance like `The actor was paid by the judge' and `The actor paid the judge'.

\noindent \textbf{Compositionality-sensitivity Datasets (IS-CS)}: \citet{nie2018analyzing} trained a softmax regression model using unigram pattern pair features across two sentences as well as unigram features in hypothesis and premise sentences to obtain the 'lexically misleading scores (LMS)' for each instance in the test sets. We use $\mathbf{CS}_{0.7}$ in their paper which denotes the subsets whose LMS are larger that 0.7. 

\subsubsection{Logical Inference Ability (LI)}\label{sec:logic_inference}
\noindent \textbf{Lexical Inference Test (LI-LI)}: A proper NLI system should recognize hypernyms and hyponyms; synonym and antonyms. We merge the “antonym” category in \citet{DBLP:conf/coling/NaikRSRN18} and \citet{GlocknerSG18} to assess the models' capability to model lexical inference.

\noindent \textbf{Text-fragment Swap Test (LI-TS)}: NLI system should also follow the first-order logic constraints \cite{wang2019if,minervini2018adversarially}. For example, if the premise sentence $s_p$ entails the hypothesis sentence $s_h$, then $s_h$ must not be contradicted by $s_p$. We then swap the two sentences in the original MultiNLI mismatched dev sets. If the gold label is `\emph{contradiction}', the corresponding label in the swapped instance remains unchanged, otherwise it becomes `\emph{non-contradicted}'.

\subsubsection{Stress Test (ST)} \label{sec:stress_test}
We also include the “word overlap” (\textbf{ST-WO}), “negation” (\textbf{ST-NE}), “length mismatch” (\textbf{ST-LM}) and “spelling errors” (\textbf{ST-SE}) in \citet{DBLP:conf/coling/NaikRSRN18}, in which ST-WO aims at detecting lexical overlap heuristics described in \citet{mccoy2019right} (IS-SD in Sec \ref{sec:inter-sentence}); ST-NE aims at detecting strong negative lexical cues in partial-input sentences like PI-SP in Sec \ref{sec:inter-sentence}.

\begin{figure}[t]
\begin{center}
\includegraphics[width=1.0\linewidth]{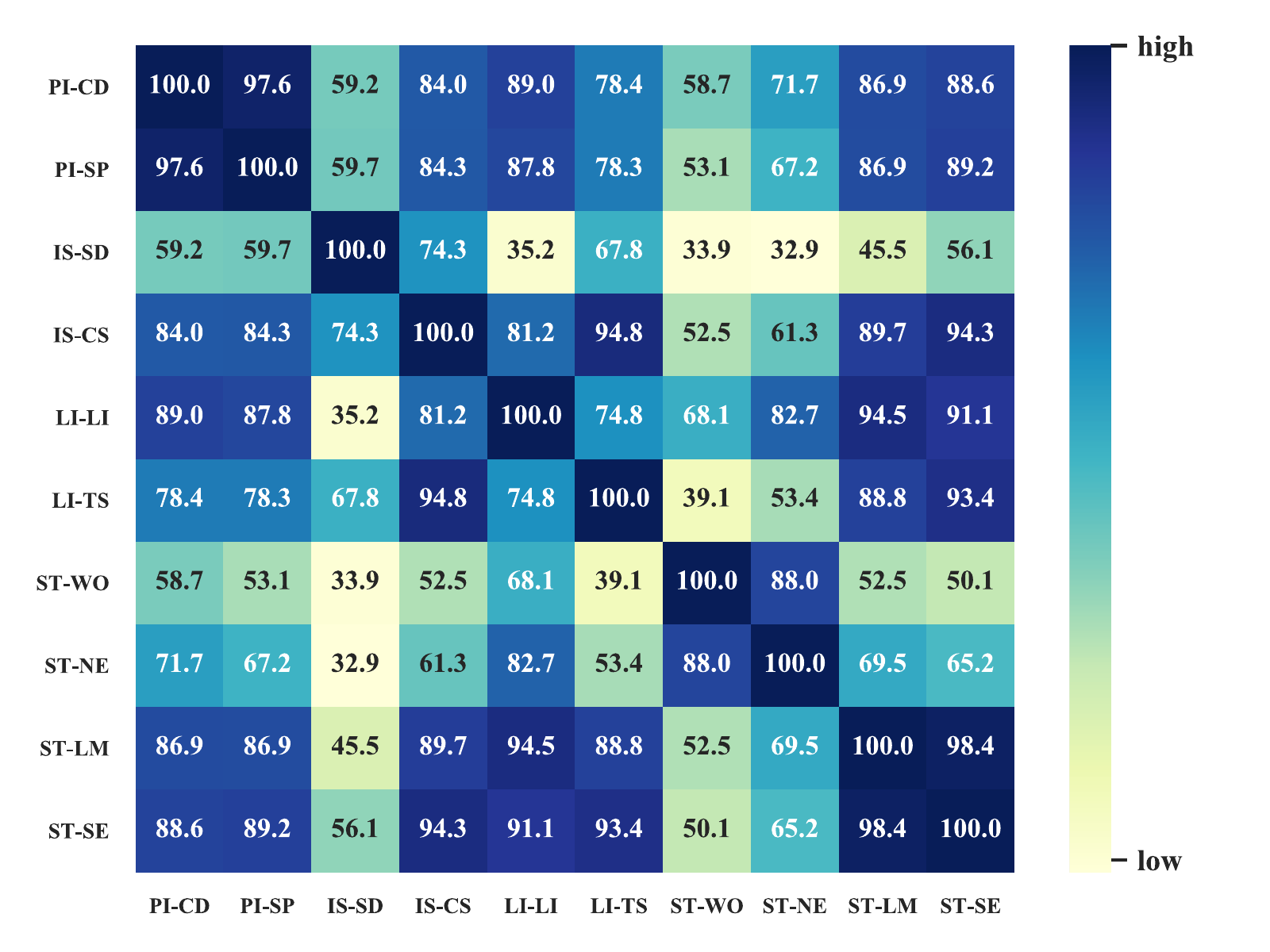}
\end{center}
\caption{The surrogate correlations between different adversarial datasets. We show the Pearson's correlation coefficients of model performance on different adversarial datasets in different runs (Sec \ref{sec:data_correlation}).}\label{fig:data_correlation}
\vspace{-0.3cm}
\end{figure}

\subsubsection{Insights within Adversarial Tests}\label{sec:data_correlation}
To provide actionable insights to NLP practitioners, we list how these adversarial instances constructed and why they might fail NLI models in Table \ref{tab:data_categorize}. 
Those adversarial datasets are potentially correlated with each other due to similar constructing process or constructing goals. 
For example, `PI-CD', `PI-SP' and `IS-CS' are all created with instance selection from original test sets in order to attack the models which improperly rely on the superficial lexical patterns, thus they might be potentially correlated. Although we could analytically assess the correlation between adversarial datasets, it is hard to demonstrate their underlying relationships from a quantitative perspective.
We instead try to utilize the model performances on these adversarial datasets as surrogates to visualize their correlations.
Concretely, we first collect the model accuracy scores on each adversarial dataset according to 30 runs of 10 baseline models (3 runs each) listed in Table \ref{tab:main_results}. Then we show the pearson correlation coefficients of the model scores on any two distinct adversarial datasets in Fig \ref{fig:data_correlation}.
According to Fig \ref{fig:data_correlation}, `IS-SD' (HANS) has higher correlation with `IS-CS' and `LI-TS' compared with other adversarial datasets, we assume this is because they are constructed based on cross sentence heuristics in the natural occurring settings, as opposed to stress test datasets which add tautology like `and true is true' to the end of hypothesis sentences \cite{DBLP:conf/coling/NaikRSRN18}. `LI-LI' instances are created by few lexical changes on premise sentence which would easily fall into `word overlap' heuristics as elaborated in the `IS-SD' dataset, thus `LI-LI' has low correlation with `IS-SD'.

\begin{table}[]
\small
    \centering
    \begin{tabular}{lcccc}
\hline
  & SNLI & MNLI & DNLI & ANLI  \\
Train &549362 &392702 &249947 &162765  \\
Valid &9842 &9832 &31696 &2200 \\
Test  &9824 &9815 &31232 &2200 \\
\hline
    \end{tabular}
    \caption{Statistics for datasets used in Sec \ref{sec:largerdata_main}. For MNLI, we utlize the matched dev and mismatched dev sets as valid and test sets respectively.}\label{tab:all4_size}
\vspace{-0.3cm}
\end{table}

\begin{table*}[t]
\small
\centering
\setlength{\tabcolsep}{1.4mm}{
\begin{tabular}{lcccccccc|ccccc|c}
\toprule
& \multicolumn{8}{c|}{Adversarial Test} & \multicolumn{5}{c|}{Generalization Power Test} & \\
& PI-CD &  PI-SP & IS-SD  & IS-CS & LI-LI & LI-TS & ST & Avg. & RTE & DIAG & SICK & SciTail & Avg. & MNLI\\ \midrule
InferSent & 52.1 & 55.3 & 53.9 & 33.5 & 43.6 & 70.5 & 53.3 & 51.7 & 61.8 & 10.6 & 25.4 & 24.7 & 30.6 & 70.5
\\+ELMO & 48.6 & 59.8 & \underline{55.2} & 42.1 & 38.5 & 72.4 & 52.7 & 52.8 & 62.5 & 9.8 & 24.6 & 18.5 & 28.9 & 72.5
\\DAM & 55.0 & 54.4 & 50.2 & 35.7 & 62.7 & 74.3 & 53.0 & 55.0 & 62.7 & 10.3 & \underline{27.0} & \underline{30.0} & 32.5 & 70.3
\\ESIM & \underline{55.1} & \underline{66.3} & 49.8 & \underline{52.7} & \underline{63.2} & \underline{79.6} & \underline{53.8} & 60.1 & \underline{66.2} & \underline{11.3} & 25.1 & 27.5 & 32.5 & \underline{77.3}
\\ \midrule
$\mathrm{BERT}_{B}$ & 72.2 & 73.9 & 63.8 & 65.4 & 85.6 & 82.6 & 63.5 & 72.4 & 75.4 & 36.2 & 54.2 & 66.1 & 58.0 & 83.5   \\
$\mathrm{BERT}_{L}$ & 74.7 & 75.5 & 70.4 & 70.6 & 87.9 & 83.8 & 67.3 & 75.7 & 77.6 & 39.4 & 55.5 & 68.3 & 60.2 & 85.7   \\
$\mathrm{XLNet}_{B}$ & 73.1 & 77.9 & 71.2 & 70.4 & 85.5 & 84.8 & 68.5 & 75.9 & 78.0 & 39.2 & 55.8 & 66.7 & 59.9 & 86.6   \\
$\mathrm{XLNet}_{L}$ & 78.8 & \underline{\textbf{81.7}} & 76.7 & 77.3 & \underline{\textbf{93.4}} & 88.5 & 72.4 & 81.3 & 83.4 & 45.9 & \underline{\textbf{57.6}} & \underline{\textbf{73.0}} & 65.0 & 89.3 \\
$\mathrm{RoBERTa}_{B}$ & 76.6 & 80.9 & 72.0 & 74.1 & 89.6 & 85.3 & 66.4 & 77.8 & 80.9 & 42.1 & 55.9 & 69.0 & 62.0 & 87.4   \\
$\mathrm{RoBERTa}_{L}$ & \underline{\textbf{80.0}} & 79.2 & \underline{\textbf{80.0}} & \underline{\textbf{77.0}} & 92.4 & \underline{\textbf{88.6}} & \underline{\textbf{73.4}} & 81.5 & \underline{\textbf{84.4}} & \underline{\textbf{50.5}} & 57.3 & 72.2 & 66.1 & \underline{\textbf{89.9}} \\ 
 \bottomrule
\end{tabular}
}
\caption{The performance of models on adversarial and generalization power tests (Sec \ref{sec:bencmark}) trained on MultiNLI. B and L in the subscript denote base and large versions of pretrained models. We use \textbf{bold} and \underline{underlined} numbers to represent the highest scores in each column/block. Same marks are also used in Table  \ref{tab:bias_section}, \ref{tab:data_aug} and \ref{tab:largerdata_rew}.}
\label{tab:main_results}
\vspace{-0.1cm}
\end{table*}

\subsection{Other Data Resources}

\noindent \textbf{Generalization Power Test}: we test the models on several general purpose datasets, including NLI diagnostic dataset (Diag) \cite{wang2018glue}, for which we use `Matthews correlation coefficient' \cite{matthews1975comparison} as the evaluation metric. We also incorporate RTE \cite{dagan2006pascal}, SICK \cite{marelli2014semeval} and SciTail \cite{khot2018scitail} in our testing.

\noindent \textbf{Training Resources}: apart from SNLI \cite{snli:emnlp2015}, and MultiNLI \cite{mnli:N18-1101}, we also incorporate Diverse NLI (DNLI) \cite{poliak2018collecting}  and Adversarial NLI (ANLI) \cite{nie2019adversarial} datasets for training.
For DNLI, we merge the subsets to form unified train/valid/test sets. Dataset Statistics are shown in Table \ref{tab:all4_size}.

\subsection{Model Performance on the Benchmark}
We show the performance of different models trained on MultiNLI in Table \ref{tab:main_results}. The general trend is that more powerful model which has higher performance on the original (in-domain) test sets (RoBERTa (large)) outperforms most models in both adversarial and general purpose settings. 

In the following sections, we investigate several model agnostic methods for debiasing NLI models. Specifically, we are interested in: 1) how to (or is it possible to) make the NLI models robust to multiple distinct adversarial attacks using a unified debiasing method and 2) how the debiasing methods influence model generalization power of NLI.

\begin{table*}[]
\small
\centering
    \begin{tabular}{lc|cc|cc|cc|ccc}
\toprule
 & Baseline & \multicolumn{2}{c|}{Word Overlap} & \multicolumn{2}{c|}{Partial Input} & \multicolumn{2}{c|}{Sentence Length} & 
\multicolumn{3}{c}{Debiasing Combination} \\
      & ($\mathrm{BERT}_{base}$) & ReW & BiasProd & ReW & BiasProd & ReW & BiasProd & MixW & AddProd & BestEn\\ \midrule
PI-CD & 72.2 & 70.9 & 71.4 & \textbf{72.6} & 71.8 & \textbf{72.6} & 72.3 & 71.9 & 71.3 & \textbf{72.6}\\
PI-SP & 73.9 & 70.6 & 70.1 & 74.7 & 73.0 & \textbf{75.2} & 73.3 & 71.7 & 70.4 & 73.9\\
IS-SD & 63.8 & 69.2 & 71.0 & 65.7 & 63.8 & 56.9 & 59.5 & 54.6 & 61.5 & \textbf{72.5}\\
IS-CS & 65.4 & 64.8 & 64.2 & 67.1 & \textbf{68.9} & 64.9 & 66.9 & 65.4 & \textbf{68.9} & 64.9\\
LI-LI & 85.6 & 87.0 & 87.8 & 86.0 & 85.0 & 85.7 & 85.5 & 86.8 & \textbf{88.4} & 87.7\\
LI-TS & 82.6 & 81.8 & 81.7 & 82.0 & 82.3 & 81.3 & 83.7 & 82.3 & 81.9 & \textbf{84.5}\\
ST-LM & 82.2 & 82.3 & 81.7 & 81.6 & 81.1 & 82.6 & 82.7 & 82.6 & 79.9 & \textbf{83.1}\\\hline
Gen. Avg. & 58.0 & 56.8 & 56.6 & 57.5 & 56.7 & 57.9 & 57.5 & 57.1 & 55.9 & \textbf{58.1}\\\hline
MNLI & 83.5 & 84.2 & 82.8 & 84.3 & 83.3 & 80.3 & 80.9 & 84.0 & 81.2 & \textbf{84.5}\\ \bottomrule
        
    \end{tabular}
    \caption{The performance of debiasing methods (Sec \ref{sec:debias_main}) based on BERT base model (baseline) trained on MultiNLI. 
    ReW, BiasProd refer to instance reweighting and bias product ensemble methods in Sec \ref{sec:debias_methods}. Word overlap, partial input and sentence length are the known biases in NLI (Sec \ref{sec:known_bias}). MixW, AddProd, BestEn are our trials to combine distinct debiasing methods (Sec \ref{sec:debias_merge}). `Gen. Avg' is the average score of test sets in generalization power test. \textbf{Bold} numbers mark the highest score (besting debiasing model) \emph{in each row}.
}
    \label{tab:bias_section}
\vspace{-0.2cm}
\end{table*}

\section{Mixture of Experts (MoE) Debiasing}\label{sec:debias_main}
We utilize the MoE ensemble model \citet{clark2019don} as the backbone to mitigate three known biases in NLI. Concretely, we implement the `instance reweighting' and `bias product' methods in \citet{clark2019don}. Based on these methods, we perform several trials on combating several distinct NLI biases at the same time. 
\subsection{Debiasing Methods}\label{sec:debias_methods}
\noindent \textbf{Notations}: for a known NLI bias, they firstly train a bias-only model $B$ and then use its output $\mathbf{b}$ as a guidance to train the prime model. 
In the context of three-way NLI training, $\mathbf{b_i}$ is a normalized 3-element vector which represents the predicted possibility of each NLI label for i-th training example. Suppose $\mathbf{p_i}$ is output of the prime model which has the same meaning as $\mathbf{b_i}$.

\noindent \textbf{Instance Reweighting}: suppose $b_i^{y_i}$ is the possibility that the bias-only model assigns to the correct label $y_i$ for $i$-th training example. They trained the models in a weighted version of the data, where the weight $\alpha_i$ for the $i$-th training example is (1-$b_i^{y_i}$). 
The loss function for a training batch with $k$ examples is a weighted sum of instance-level loss $l_i$: $\mathbf{L}_{batch} = \alpha_i * l_i / (\sum_{i=1}^k \alpha_i)$.

\noindent \textbf{Bias Product Ensemble}: an ensemble method that is a product of experts $\mathbf{\hat{p_{i}}} = softmax(log(\mathbf{p_i})+log(\mathbf{b_i}))$.    

By doing so, the prime model would be encouraged to learn all the information except the specific bias. An intuitive justification from the probabilistic view can be found in \citet{clark2019don}. Note that while training, only the prime model is updated while the bias-only model remains unchanged.

\subsection{Known Biases in NLI}\label{sec:known_bias}
\noindent \textbf{Word overlap heuristics}: To combat the word overlap heuristics (HANS \cite{mccoy2019right}, renamed as IS-SD in Sec \ref{sec:inter-sentence}), \citet{clark2019don} used the following features to train a bias-only model: (1) whether the hypothesis is a sub-sequence of the premise, (2) whether all words in the hypothesis appear in the premise, (3) the percent of words from the hypothesis that appear in the premise, (4) the average and the max of the minimum distance between each premise word with each hypothesis word. We use their trained bias-only model output for experiments.

\noindent \textbf{Partial input heuristics}: To combat the hypothesis-only bias in NLI (PI-CD and PI-SP in Sec \ref{sec:partial-input}), we use RoBERTa (base) model to train a bias-only model by taking only hypothesis sentences as inputs. Our hypothesis-only model gets 60.4\% accuracy on the mismatched dev set of MultiNLI, which is higher than the reported numbers in \citet{gururangan2018annotation} (52.3\%) and \citet{poliak2018hypothesis} (55.18\%).

\noindent \textbf{Sentence length heuristics}: \citet{gururangan2018annotation} shows that the length of hypothesis and premise over different labels is not evenly distributed (ST-LM in Sec \ref{sec:stress_test}). So we trained a bias-only classifier based on the following sentence length related features: 1) the sentence lengths of hypothesis and premise sentences, 2) the mean and difference of these lengths. Our classifier achieves 41.3\% accuracy on the mismatched dev set of MultiNLI, which outperforms the majority class baseline by 6.1\%.

\begin{table*}[]
\small
\centering
\setlength{\tabcolsep}{1.2mm}{
\begin{tabular}{lcccccccc|ccccc|c}
\toprule
& \multicolumn{8}{c|}{Adversarial Test} & \multicolumn{5}{c|}{Generalization Power Test} & \\
& PI-CD &  PI-SP & IS-SD  & IS-CS & LI-LI & LI-TS & ST & Avg. & RTE & DIAG & SICK & SciTail & Avg. & MNLI\\ \midrule
Baseline & \textbf{72.2} & 73.9 & 63.8 & 65.4 & 85.6 & 82.6 & 63.5 & 72.4 & 75.4 & \textbf{36.2} & 54.2 & 66.1 & \textbf{58.0} & 83.5   \\
Text Swap & 71.7 & 72.8 & 63.5 & \textbf{67.4} & \textbf{86.3} & \textbf{86.8} & \textbf{66.5} & \textbf{73.6} & 73.3 & 35.3 & 54.7 & \textbf{66.8} & 57.6 & \textbf{83.7}\\
Sub (synonym) & 69.8 & 72.0 & 62.4 & 65.8 & 85.2 & 82.8 & 64.3 & 71.8 & 74.4 & 34.2 & 55.1 & 65.8 & 57.4 & 83.5 \\
Sub (MLM) & 71.0 & 72.8 & 64.4 & 65.9 & 85.6 & 83.3 & 64.9 & 72.6 & 74.8 & 34.7 & \textbf{55.4} & 65.7 & 57.7 & 83.6 \\
Paraphrase & 72.1 & \textbf{74.6} & \textbf{66.5} & 66.4 & 85.7 & 83.1 & 64.8 & 73.3 & \textbf{75.8} & 35.1 & 55.0 & 65.0 & 57.7 & \textbf{83.7}
\\\bottomrule
\end{tabular}
}
\caption{The performance of BERT base model under different data augmentation strategies (Sec \ref{sec:dataaug_main}). }
\label{tab:data_aug}
\vspace{-0.3cm}
\end{table*}

\subsection{Combating Distinct Biases}\label{sec:debias_merge}
Suppose we already have $m$ bias-only models $\{B^1,B^2,\cdots,B^m\}$ and the corresponding output $\{\mathbf{b^1},\mathbf{b^2},\cdots,\mathbf{b^m}\}$ at hand, we test three different approaches to integrate these models.

\noindent \textbf{MixWeight}: Using the product of weights from different debiasing models while performing instance reweighting. We replace the weight for the $i$-th training example ($\alpha_i$ in Sec \ref{sec:debias_methods}) with $\prod_{j=1}^m (1-b_{i}^{y_i})$ and utilize the same loss function as `instance reweighting' in Sec \ref{sec:debias_methods}).

\noindent \textbf{AddProduct}: We view different bias-only models as multiple independent experts and then apply the bias product ensemble as `bias product ensemble' in Sec \ref{sec:known_bias}: $\mathbf{\hat{p_{i}}} = softmax(log(\mathbf{p_i})+ \sum_{j=1}^m log(\mathbf{b_i^j}))$.

\noindent \textbf{BestEnsemble}: We also try to ensemble the best single debiasing models. In our experiments (Table \ref{tab:bias_section}), we ensemble the three reweighting models (`ReW' models in column 2,4 and 6) for each bias to form the BestEnsemble model.

\subsection{Discussions for MoE Methods}
For \textbf{mixture of experts model}, we summarize our findings from Table \ref{tab:bias_section} below: 

\noindent1)  
For all three known biases in Sec \ref{sec:known_bias}, we find that the debiasing methods targeting at specific known biases increase the model performance on the corresponding adversarial datasets, e.g. for the word overlap heuristics, BiasProd model gets 71.0\% accuracy on IS-SD (HANS) test set, 7.2\% higher than baseline.

\noindent2) The bias-specific methods might not make the NLI models more robust and generalized. For example, the methods designed for word overlap heuristics get lower scores on PI-CD, PI-SP, IC-CS, LI-TS test sets than the baseline model.

\noindent3) The proposed debiasing merging methods BestEn (Sec \ref{sec:debias_merge}) inherits the advantages of the 4 bias-specific methods on PI-CD, IS-SD, LI-TS and ST-LM compared with other MoE debiasing models.


\section{Data Augmentation} \label{sec:dataaug_main}
In this section, we explore 3 automatic augmentation ways without collecting new data. For fair comparison, in all the following settings, we double the training size by automatically generating the same number of augmented instances as the original training sets as shown in Table \ref{tab:data_aug}.  
\subsection{Methods}
\noindent \textbf{Text Swap}: It is an easy-to-implement method which swaps the premise $p$ and hypothesis $h$ sentences in the original datasets. It might be an potential solution to combat the partial-input heuristics (Sec \ref{sec:partial-input}) as the superficial patterns are not observed in the premise sentences.
According to the first-order logic rules (LI-TS in Sec \ref{sec:logic_inference}), we can only determine the gold labels for the swapped sentence pairs whose original labels are \emph{contradiction}. For the \emph{entailment} and \emph{neutral} instances, we using the ensembled RoBERTa large model trained on `all4' training set (Table \ref{tab:largerdata_rew}) to label the swapped sentence pairs.

\noindent \textbf{Word Substitution}:
We also tried to create new training instances by flipping the words in the hypothesis sentences.
We try two ways to perform substitution: 1) \textbf{synonym}:  We use NLTK \cite{loper2002nltk} to firstly find the synonym candidates of the content words (including nouns, verbs and adjectives) in the hypothesis sentences, and then we replace the content words with their synonyms if the cosine similarity ([-1,1]) between the original window and the window after replacement is larger than 0. The window contains at most 3 words including the replaced word and its neighbours. We represent that window by max-pooling over the 300d Glove \cite{pennington2014glove} embedding of the words in that window.
2) \textbf{Masked LM}: we randomly select 30\% content words and then load the pretrained BERT large model to perform masked LM task. We uniformly sample from top-100 ranking candidate words (excluding the original word) and then replace the original content word with the sampled one.

\begin{table*}[]
    \centering
\small
\centering
\setlength{\tabcolsep}{0.66mm}{
\begin{tabular}{lcccccccc|ccccc|cccc}
\toprule
& \multicolumn{8}{c|}{Adversarial Test} & \multicolumn{5}{c|}{Generalization Power Test} & \multicolumn{4}{c}{Original Test Sets}\\
& PI-CD &  PI-SP & IS-SD  & IS-CS & LI-LI & LI-TS & ST & Avg. & RTE & DIAG & SICK & SciTail & Avg. & DNLI & ANLI& SNLI & MNLI\\ \midrule
\multicolumn{18}{c}{\textbf{RoBERTa (base) Model }} \\
D(only) & 38.5 & 48.2 & 55.6 & 40.9 & 12.6 & 72.9 & 40.9 & 44.2 & 54.9 & 9.1 & 40.9 & 39.4 & 36.1 & 92.9 & 32.6 & 42.1 & 47.0
\\
A(only) & 64.6 & 60.6 & 57.9 & 66.9 & 92.6 & 80.8 & 68.1 & 70.2 & 80.6 & 33.8 & 51.2 & 63.7 & 57.3 & 58.9 & 49.1 & 73.6 & 78.5
\\
S(only) & 82.2 & 64.4 & 67.4 & 62.2 & 93.2 & 80.7 & 64.6 & 73.5 & 72.5 & 36.0 & \underline{57.8} & 49.6 & 54.0 & 58.8 & 31.3 & 91.3 & 79.9
\\
M(only) & 76.6 & 80.9 & 72.0 & 74.1 & 89.6 & 85.3 & 66.4 & 77.8 & 80.9 & 42.1 & 55.9 & 69.0 & 62.0 & 59.3 & 29.4 & 84.2 & 87.4
\\
M+S & \underline{82.8} & 80.1 & 73.3 & 74.4 & 91.8 & 85.6 & 67.8 & 79.4 & 81.2 & 40.7 & 57.5 & 67.4 & 61.7 & 60.5 & 28.3 & 91.7 & 87.4
\\
M+S+D & 82.7 & 79.8 & 75.1 & 72.9 & 92.1 & 84.7 & 68.1 & 79.3 & 80.4 & 40.9 & 57.1 & 68.3 & 61.8 & 92.8 & 30.3 & 91.7 & 87.7
\\
All4 & 82.6 & \underline{81.7} & \underline{77.0} & \underline{74.7} & 94.7 & 85.3 & \underline{69.1} & 80.7 & 83.7 & 41.9 & 57.3 & \underline{70.5} & 63.4 & \underline{93.0} & 49.2 & \underline{91.9} & 87.7
\\ 
All4+SR & 82.6 & 82.5 & 74.7 & 73.8 & \underline{95.2} & \underline{86.0} & 69.0 & 80.5 & \underline{83.9} & 41.3 & 57.3 & 69.6 & 63.0 & 92.8 & 49.1 & 91.7 & \underline{87.8}
\\
All4+PR &83.4 & 79.5 & 75.5 & 73.8 & 94.6 & 85.5 & \underline{69.1} & 80.2 & 83.8 & \underline{44.0} & 57.5 & \underline{70.5} & 64.0 & 92.9 & \underline{51.2} & \underline{91.9} & 87.6
\\\midrule
\multicolumn{18}{c}{\textbf{RoBERTa (large) Model }}
\\
All4& 84.6 & \underline{\textbf{83.8}} & 79.6 & \underline{\textbf{79.3}} & 94.9 & 88.6 & 71.6 & 83.2 & 87.6 & \underline{\textbf{50.2}} & 57.9 & 73.1 & 67.2 & 93.2 & 55.5 & 92.7 & 90.4
\\
All4+ME & \underline{\textbf{85.0}} & 81.4 & \underline{\textbf{80.1}} & 77.7 & \underline{\textbf{95.7}} & 88.7 & 72.2 & 83.0 & 87.2 & 47.4 & \underline{\textbf{58.0}} & 73.7 & 66.6 & \underline{\textbf{93.3}} & 54.8 & \underline{\textbf{93.0}} & 90.2
\\
All4+SE & \underline{\textbf{85.0}} & 81.9 & 77.5 & 77.9 & 95.4 & \underline{\textbf{89.2}} & \underline{\textbf{72.5}} & 82.8 & \underline{\textbf{88.5}} & 49.3 & 57.9 & \underline{\textbf{73.9}} & 67.4 & \underline{\textbf{93.3}} & \underline{\textbf{55.7}} & \underline{\textbf{93.0}} & \underline{\textbf{90.6}}
\\\bottomrule
\end{tabular}
}
\caption{Performance of RoBERTa model trained on different datasets using multiple reweighting and ensemble strategries (Sec \ref{sec:largerdata_main}). `D', `A', `S', `M', `All4' denotes DNLI, ANLI, SNLI, MNLI and the merge of all 4 datasets respectively. `M+S' is created by merging MNLI and SNLI datasets, same principle in other settings. `ME' and `SE' denote the ensemble strategies in Sec \ref{sec:largerdata_ensemble}: the ensemble of 3 distinct models: BERT(large), XLNet(large) and RoBERTa(large) and the ensemble of 3 RoBERTa(large) models. `SR' and `PR' refer to the size-based and performance-based reweighting in Sec \ref{sec:largerdata_reweight}. Here for `PR' we use the average score of all the listed tests in `D(only)', `A(only)', `S(only)' and `M(only)' rows as their corresponding performance. }
\label{tab:largerdata_rew}
\vspace{-0.3cm}
\end{table*}

\noindent \textbf{Paraphrase}:
We create the paraphrases for the original hypothesis sentences by back translation \cite{wieting2017paranmt,hu2019improved} using the pretrained English-German and German-English machine translation models \cite{ng2019facebook}. To increase the diversity, we use beam search (size=5) for German-English translation and get the paraphrase by sampling from the candidate sentences.

\subsection{Quality Analysis}
To assess the quality of augmented data, we conduct both automatic and human evaluation. For automatic evaluation, we use the best NLI model (RoBERTa(large) model with `All4+SinEN' in Table \ref{tab:largerdata_rew}) in this paper to judge if the labels of augmented data are consistent with the predictions of our best NLI model. 
For human evaluation, we firstly sample 50 instances from each augmented training data and then hire 3 human annotators to decide the relation for the sentences pairs. We shuffle the 200 instances without showing the annotators the augmentation method for certain instances. We also ask the annotators to be objective and not to guess the augmentation methods and then use the majority vote for final annotation. The accuracy of text swap, word substitution (synonym), word substitution (MLM) and paraphrase are 84.0\%, 82.0\%, 88.1\% and 92.9\% respectively based on human-annotated gold labels. Correspondingly, word substitution (synonym), word substitution (MLM) and paraphrase get 76.9 \%, 83.5\% and 94.5\% accuracy on the automatic evaluation. Paraphrase augmentation is shown to have the highest quality among the four methods.

\subsection{Discussions for Data Augmentations}
For \textbf{Data Augmentation},
we show the performance of a BERT base model using different data augmentation methods in Table \ref{tab:data_aug}.

Text swap method increases the model performance on IS-CS, LI-LI, LI-TS and ST test sets, as it can make the data distribution in the premises and hypotheses more balanced. It is also an easy-to-implement method which could serve as a baseline to evaluate other automatic data augmentation methods.
For the other two methods,
the fragility of NLI models to partial input and inter-sentence heuristics is partially due to the rigid word-label concurrence (PI-SP in Sec \ref{sec:partial-input}) or word-to-word mapping (IS-SD, IS-CS in Sec \ref{sec:inter-sentence}).
More diverse lexical choices via word substitution or paraphrase might help to relieve the biases caused by these heuristics. We see that `word sub' in Table \ref{tab:data_aug} outperforms baseline on IS-CS, LI-TS and ST; `paraphrase' outperforms the baseline on IS-SD, LI-TS.
However, these two methods get lower scores on other adversarial and general purpose datasets as these debiasing techniques bias the model towards being robust to a specific bias, so it compensates by trading off performance.

\section{Dataset Merging and Model Ensemble}\label{sec:largerdata_main}
In this section we explore 1) to what extend larger dataset and ensemble would make the NLI models more robust to distinct adverserial datasets. 2) what is the best way to combine the large-scale NLI training sets in very different domains.
\subsection{Merging Heterogeneous Datasets}\label{sec:largerdata_reweight}
To set up more diverse and stronger baselines for the proposed benchmark datasets, we use 4 large-scale training datasets: SNLI, MNLI, DNLI and ANLI for the following experiments.
Those training sets are created using different strategies. Specifically, SNLI and MNLI are created in a human elicited way \cite{poliak2018hypothesis}: the human annotators are asked to write a hypothesis sentence according to the given premise and label. DNLI recasts other NLP tasks to fit in the form of NLI. 
ANLI is created as hard datasets that may fail the models.  
Since those datasets vary in sizes, domains and collection processes, they might have different contribution to the final predictions. Here we investigate two instance reweighting methods accordingly.

\noindent \textbf{Notations}: suppose we have $k$ training sets $\{T_i\}_{i=1}^k$ whose sizes are $\{n_i\}_{i=1}^k$. The accuracies of a baseline model trained on $\{T_i\}_{i=1}^k$ are $\{p_i\}_{i=1}^k$ respectively. $p_i$ can be the average scores of multiple test sets or the score on an single in-domain/ out-of-domain/ adversarial test set.

\noindent \textbf{Size-based reweighting (SR)}: Smaller training sets might have less influence on the models than larger ones. In this setting, we try to increase the weight of smaller datasets so that each dataset contributes more equally to the final predictions. 
We implement this reweighting method by replacing the $\alpha_i$ in Sec \ref{sec:debias_methods} with $(\sum_k n_k) / n_i (i \in T_i)$.

\noindent \textbf{Performance-based reweighting (PR)}: Different training sets may vary in annotation quality and collection process thus have distinct model performance. In this setting, we reweight the training instances with the performance of a baseline model on the specific training sets. We still use the instance weights in Sec \ref{sec:debias_methods} with $\alpha_i =  p_i / (\sum_k p_k) (i \in T_i)$.

\subsection{Model Ensemble} \label{sec:largerdata_ensemble}
We try two modes for model ensemble: \textbf{mixed} and \textbf{single} mode. In the mixed mode, we ensemble three different models (BERT, XLNet, RoBERTa) while in the single mode, we ensemble three same models (RoBERTa*3). 

\subsection{Discussions}
For \textbf{Dataset merging and model ensemble},
according to Table \ref{tab:largerdata_rew},
We find that:

\noindent 1) Incorporating heterogeneous training data is a straightforward method to enhance the robustness of NLI models. Empirically we see incorporating datasets with adversarial human-in-the-loop annotating (e.g. ANLI) is more efficient that incorporating automatically constructed dataset without human curation (e.g. DNLI).

\noindent 2) In RoBERTa base model, the `All4+PR' model get higher scores on diagnostic and ANLI test sets than `All4' baseline, which shows that increasing the weight of higher quality dataset may help to increase accuracy on certain test sets. Notably, performance based reweighting helps the model gain 2 points (49.2 vs 51.2) on ANLI compared with baseline model while keeping the inference ability on DNLI, SNLI and MNLI test sets.

\noindent 3) In RoBERTa large model, we see that on some datasets, like IS-SD, the mixed ensemble model may even outperform the single ensemble model even if its two components (XLNet and BERT) are less powerful than those (RoBERTa) in single ensemble mode.

\begin{figure}[t]
\begin{center}
\includegraphics[width=1.0\linewidth]{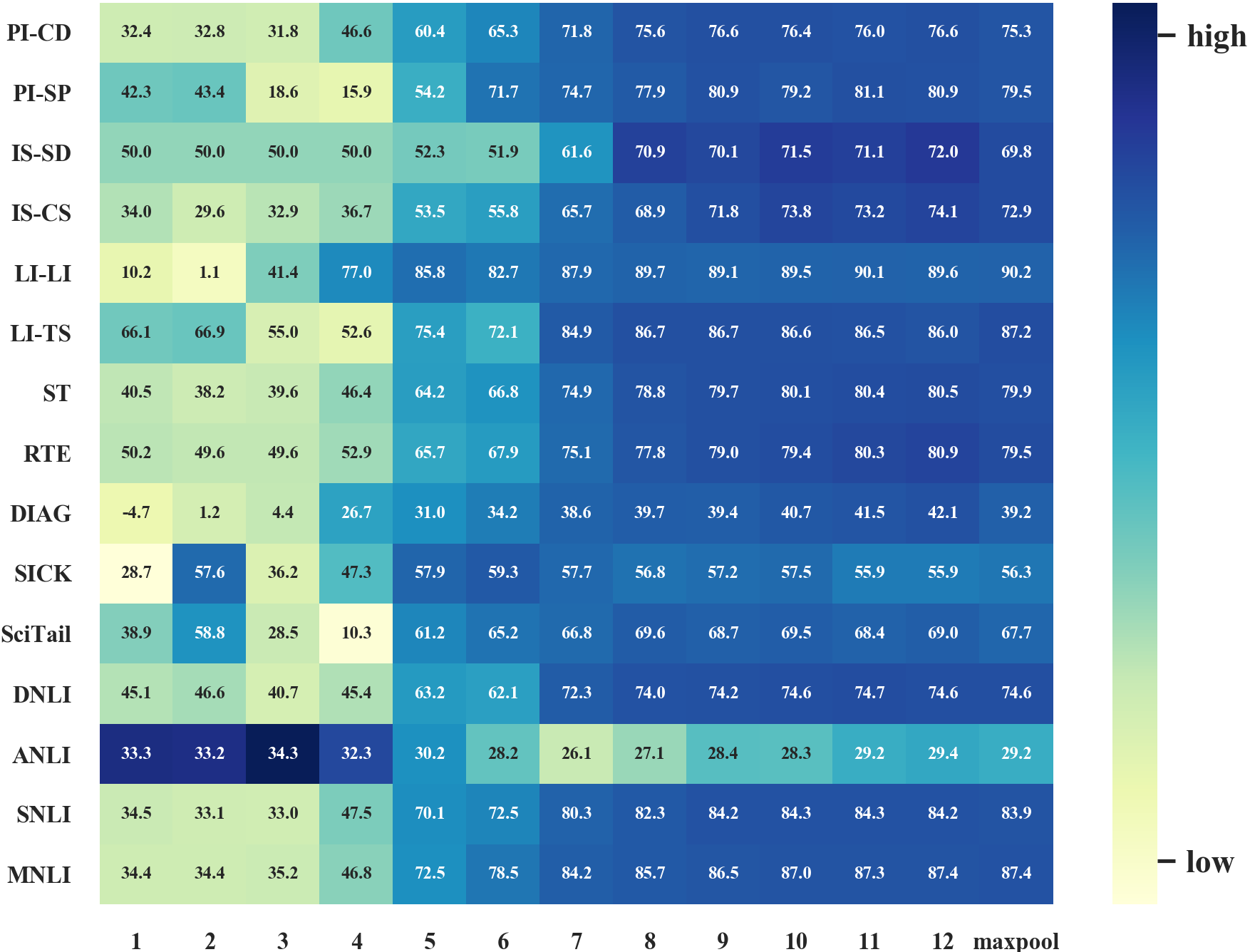}
\end{center}
\caption{Per-layer analysis for RoBERTa(base) model trained on MultiNLI. Darker blue denotes higher score. `max' represents the maxpooled vector across all layers.
Nearly all test sets except ANLI get higher scores by using higher layers. On ANLI, the performance of the first 4 layers are close to random guess while that of higher layers is about 4 point lower than random guess.}\label{fig:per_layer_analysis}
\end{figure}

\begin{table}[]
\centering
\small
\setlength{\tabcolsep}{1mm}{
\begin{tabular}{|c|c|c|}
\hline
Labels & Transformation & Datasets \\
    ($\neg$E, E) & C$\Rightarrow\neg$E, N$\Rightarrow\neg$E & IS-SD, RTE, DNLI \\
    ($\neg$C, C) & E$\Rightarrow\neg$C, N$\Rightarrow\neg$C & LI-TS\\
    (E, C) & - & LI-LI\\
    (N, E) & - & SciTail \\\hline
    \end{tabular}}
\caption{How we evaluate the test sets with only two labels in 3-way NLI classification. E,C,N,$\neg$ means \emph{entailment}, \emph{contradiction}, \emph{neutral} and \emph{not} respectively. $\Rightarrow$ means changing the left-hand side model prediction with the right-hand side label while evaluation.}
\label{tab:my_label}
\end{table}

\section{Experimental Settings}
\subsection{Implementation Details}
We set up both pretrained and non-pretrained model baselines for the proposed evaluation bechmarks. We rerun their public available codebases \cite{wolf1910huggingface}, including InferSent \cite{conneau2017supervised} \footnote{\url{https://github.com/facebookresearch/InferSent}} (w/ and w/o Elmo \cite{peters2018deep}), DAM \cite{parikh2016decomposable} \footnote{\url{https://github.com/harvardnlp/decomp-attn}}, ESIM \cite{chen2016enhanced}\footnote{\url{https://github.com/coetaur0/ESIM}}, BERT (uncased) \cite{devlin2018bert}, XLNet (cased) \cite{yang2019xlnet} and RoBERTa \cite{liu2019roberta}, \footnote{\url{https://github.com/huggingface/transformers}}.
we map the vector at the position of the `[CLS]' token in the pretrained models to three-way NLI classification via linear transformation. We show the per-layer analyses for RoBERTa model in Table \ref{fig:per_layer_analysis}.
We try to reduce the randomness of our experiments by 3 runs using different random seeds. We report the median of the 3 runs for all the tables except the ensemble-related (Sec \ref{sec:largerdata_ensemble}) experiments in Table \ref{tab:largerdata_rew}.
Table \ref{tab:my_label} shows how we evaluate the test sets with only two labels in 3-way NLI classification.

\begin{table}[]
\small
\centering
\setlength{\tabcolsep}{1mm}{
\begin{tabular}{lcccccc|c}
\toprule
 & RTE & SICK & SciTail & DNLI & ANLI& SNLI & MNLI\\ \midrule
Origin & 75.4 & 54.2 & 66.1 & 54.2 & 27.7 & 80.0 & \textbf{83.5}
\\
Mixed & \textbf{75.5} & 54.3 & \textbf{67.3} & 54.8 & 27.4 & 79.9 & 83.4\\
Oracle & \textbf{75.5} & \textbf{55.2} & 
\textbf{67.3} & \textbf{56.7} & \textbf{28.0} & \textbf{80.3} & \textbf{83.5}\\
 \bottomrule
\end{tabular}
}
\caption{The performance of BERT base model under different model selection strategies.}
\label{tab:model_selection}
\end{table}

\subsection{Model Selection Strategy}
Since we test the NLI models on multiple general-purpose dataset. it is an important question how we choose the dev set. We explore 3 different model selection settings: 

\noindent 1) \textbf{Origin}: using the original in-domain dev set. 

\noindent 2) \textbf{Mixed}: using the merged dev sets which include all the instances in the in-domain and extra dev sets in generalization power tests. 

\noindent 3) \textbf{Oracle}: tuning the model for each generalization power test using its own dev set.

We show the performance of a BERT base model trained on MultiNLI utilizing the above mentioned model selection strategies in Table \ref{tab:model_selection}. In this paper we use the `origin' mode, as it is too expensive to use the `oracle' strategy in all experiments, besides we did not see much difference between the `mixed' and `origin' modes. Notably when we merge different training sets, we also merge their dev sets correspondingly to form a unified in-domain dev set in Table \ref{tab:largerdata_rew}.

\section{Related Work}
\noindent \textbf{Bias in NLI}: The bias in the data annotation exists in many tasks, e.g. lexical inference \cite{levy2015supervised}, visual question answering \cite{goyal2017making}, ROC story cloze \cite{DBLP:conf/acl/CaiTG17,DBLP:conf/conll/SchwartzSKZCS17} etc.
The NLI models are shown to be sensitive to the compositional features in premises and hypotheses\cite{nie2018analyzing,dasgupta2018evaluating}, data permutations \cite{schluter2018data,wang2019if} and vulnerable to adversarial examples \cite{iyyer2018adversarial,minervini2018adversarially,GlocknerSG18} and crafted stress test \cite{geiger2018stress,DBLP:conf/coling/NaikRSRN18}. 
Other evidences of artifacts include sentence occurrence \cite{zhang2019selection}, syntactic heuristics between hypotheses and premises \cite{mccoy2019right} and black-box clues derived from neural models \cite{gururangan2018annotation,poliak2018hypothesis,he2019unlearn}.
\citet{RudingerMD17} showed hypotheses in SNLI has the evidence of gender, racial stereotypes, etc. 
\citet{sanchez2018behavior} analysed the behaviour of NLI models and the factors to be more robust.
\citet{feng2019misleading} discussed how to use partial-input baseline in future dataset creation.
\citet{belinkov2019adversarial,clark2019don,he2019unlearn,yaghoobzadeh2019robust,ding2020generative,utama-etal-2020-mind} proposed efficient methods to mitigate a particular known bias in NLI.
\citet{DBLP:journals/corr/abs-2010-03338,DBLP:journals/corr/abs-2009-12303} investigates the method to improve model generalization by debiasing unknown dataset bias in natural language understanding and question answering tasks.

\noindent \textbf{Benchmark collection in NLI}: GLUE \cite{wang2018glue,wang2019superglue} benchmark contains several NLI-related benchmark datasets. However it does not include adversarial test sets, domain specific test \cite{romanov2018lessons,ravichander2019equate}.  
Researchers create NLI datasets using different collection criteria, such as recasting other NLP tasks to NLI \cite{poliak2018collecting}, iteratively filtering adversarial training data by model decisions \cite{le2019adversarial} (model-in-the-loop),  counterfactually augmenting training data by human editing examples to break the model \cite{kaushik2019learning} (human-in-the-loop) and multi-round annotating depending on both human and model decisions \cite{nie2019adversarial}. 



\section{Conclusions}
We try to investigate how to build robust and generalized NLI models by model-agnostic debiasing strategies, including mixture of experts ensemble (MoE), data augmentation (DA), dataset merging and model ensemble, and benchmark these methods on various adversarial and general purpose datasets. Our findings suggest model-level MoE ensemble, text swap DA and performance based dataset merging would effectively combat multiple (though not all) distinct biases.

Although we haven't found a debiasing strategy that can guarantee the NLI models to be more robust on every adversarial dataset used in this paper, we leave the question of whether such a debiasing method exists for future research.

\section*{Acknowledgments}
We would like to thank Sam Wiseman and Kevin Gimpel for very thoughtful discussions, and the anonymous reviewers for their helpful feedback. This project is supported by NSFC (No. 61876004, No. U19A2065) and Beijing Academy of Artificial Intelligence (BAAI).

\bibliography{acl2019}

\begin{thebibliography}{58}
\expandafter\ifx\csname natexlab\endcsname\relax\def\natexlab#1{#1}\fi

\bibitem[{Belinkov et~al.(2019)Belinkov, Poliak, Shieber, Van~Durme, and
  Rush}]{belinkov2019adversarial}
Yonatan Belinkov, Adam Poliak, Stuart Shieber, Benjamin Van~Durme, and
  Alexander Rush. 2019.
\newblock \href {https://doi.org/10.18653/v1/S19-1028} {On adversarial removal
  of hypothesis-only bias in natural language inference}.
\newblock pages 256--262, Minneapolis, Minnesota. Association for Computational
  Linguistics.

\bibitem[{Bird and Loper(2004)}]{loper2002nltk}
Steven Bird and Edward Loper. 2004.
\newblock \href {https://www.aclweb.org/anthology/P04-3031} {{NLTK}: The
  natural language toolkit}.
\newblock In \emph{Proceedings of the {ACL} Interactive Poster and
  Demonstration Sessions}, pages 214--217, Barcelona, Spain. Association for
  Computational Linguistics.

\bibitem[{Bowman et~al.(2015)Bowman, Angeli, Potts, and
  Manning}]{snli:emnlp2015}
Samuel~R. Bowman, Gabor Angeli, Christopher Potts, and Christopher~D. Manning.
  2015.
\newblock \href {https://doi.org/10.18653/v1/D15-1075} {A large annotated
  corpus for learning natural language inference}.
\newblock In \emph{Proceedings of the 2015 Conference on Empirical Methods in
  Natural Language Processing}, pages 632--642, Lisbon, Portugal. Association
  for Computational Linguistics.

\bibitem[{Bras et~al.(2020)Bras, Swayamdipta, Bhagavatula, Zellers, Peters,
  Sabharwal, and Choi}]{le2019adversarial}
Ronan~Le Bras, Swabha Swayamdipta, Chandra Bhagavatula, Rowan Zellers,
  Matthew~E. Peters, Ashish Sabharwal, and Yejin Choi. 2020.
\newblock \href {http://arxiv.org/abs/2002.04108} {Adversarial filters of
  dataset biases}.
\newblock \emph{CoRR}, abs/2002.04108.

\bibitem[{Cai et~al.(2017)Cai, Tu, and Gimpel}]{DBLP:conf/acl/CaiTG17}
Zheng Cai, Lifu Tu, and Kevin Gimpel. 2017.
\newblock \href {https://doi.org/10.18653/v1/P17-2097} {Pay attention to the
  ending:strong neural baselines for the {ROC} story cloze task}.
\newblock In \emph{Proceedings of the 55th Annual Meeting of the Association
  for Computational Linguistics (Volume 2: Short Papers)}, pages 616--622,
  Vancouver, Canada. Association for Computational Linguistics.

\bibitem[{Chen et~al.(2017)Chen, Zhu, Ling, Wei, Jiang, and
  Inkpen}]{chen2016enhanced}
Qian Chen, Xiaodan Zhu, Zhen-Hua Ling, Si~Wei, Hui Jiang, and Diana Inkpen.
  2017.
\newblock \href {https://doi.org/10.18653/v1/P17-1152} {Enhanced {LSTM} for
  natural language inference}.
\newblock In \emph{Proceedings of the 55th Annual Meeting of the Association
  for Computational Linguistics (Volume 1: Long Papers)}, pages 1657--1668,
  Vancouver, Canada. Association for Computational Linguistics.

\bibitem[{Clark et~al.(2019)Clark, Yatskar, and Zettlemoyer}]{clark2019don}
Christopher Clark, Mark Yatskar, and Luke Zettlemoyer. 2019.
\newblock \href {https://doi.org/10.18653/v1/D19-1418} {Don{'}t take the easy
  way out: Ensemble based methods for avoiding known dataset biases}.
\newblock In \emph{Proceedings of the 2019 Conference on Empirical Methods in
  Natural Language Processing and the 9th International Joint Conference on
  Natural Language Processing (EMNLP-IJCNLP)}, pages 4069--4082, Hong Kong,
  China. Association for Computational Linguistics.

\bibitem[{Conneau et~al.(2017)Conneau, Kiela, Schwenk, Barrault, and
  Bordes}]{conneau2017supervised}
Alexis Conneau, Douwe Kiela, Holger Schwenk, Lo{\"\i}c Barrault, and Antoine
  Bordes. 2017.
\newblock \href {https://doi.org/10.18653/v1/D17-1070} {Supervised learning of
  universal sentence representations from natural language inference data}.
\newblock In \emph{Proceedings of the 2017 Conference on Empirical Methods in
  Natural Language Processing}, pages 670--680, Copenhagen, Denmark.
  Association for Computational Linguistics.

\bibitem[{Dagan et~al.(2005)Dagan, Glickman, and Magnini}]{dagan2006pascal}
Ido Dagan, Oren Glickman, and Bernardo Magnini. 2005.
\newblock \href {https://doi.org/10.1007/11736790\_9} {The {PASCAL} recognising
  textual entailment challenge}.
\newblock In \emph{Machine Learning Challenges, Evaluating Predictive
  Uncertainty, Visual Object Classification and Recognizing Textual Entailment,
  First {PASCAL} Machine Learning Challenges Workshop, {MLCW} 2005,
  Southampton, UK, April 11-13, 2005, Revised Selected Papers}, volume 3944 of
  \emph{Lecture Notes in Computer Science}, pages 177--190. Springer.

\bibitem[{Dagan et~al.(2013)Dagan, Roth, Sammons, and
  Zanzotto}]{DBLP:series/synthesis/2013Dagan}
Ido Dagan, Dan Roth, Mark Sammons, and Fabio~Massimo Zanzotto. 2013.
\newblock \href {https://doi.org/10.2200/S00509ED1V01Y201305HLT023}
  {\emph{Recognizing Textual Entailment: Models and Applications}}.
\newblock Synthesis Lectures on Human Language Technologies. Morgan {\&}
  Claypool Publishers.

\bibitem[{Dasgupta et~al.(2018)Dasgupta, Guo, Stuhlm{\"{u}}ller, Gershman, and
  Goodman}]{dasgupta2018evaluating}
Ishita Dasgupta, Demi Guo, Andreas Stuhlm{\"{u}}ller, Samuel Gershman, and
  Noah~D. Goodman. 2018.
\newblock \href {https://mindmodeling.org/cogsci2018/papers/0307/index.html}
  {Evaluating compositionality in sentence embeddings}.
\newblock In \emph{Proceedings of the 40th Annual Meeting of the Cognitive
  Science Society, CogSci 2018, Madison, WI, USA, July 25-28, 2018}.
  cognitivesciencesociety.org.

\bibitem[{Devlin et~al.(2019)Devlin, Chang, Lee, and
  Toutanova}]{devlin2018bert}
Jacob Devlin, Ming-Wei Chang, Kenton Lee, and Kristina Toutanova. 2019.
\newblock \href {https://doi.org/10.18653/v1/N19-1423} {{BERT}: Pre-training of
  deep bidirectional transformers for language understanding}.
\newblock In \emph{Proceedings of the 2019 Conference of the North {A}merican
  Chapter of the Association for Computational Linguistics: Human Language
  Technologies, Volume 1 (Long and Short Papers)}, pages 4171--4186,
  Minneapolis, Minnesota. Association for Computational Linguistics.

\bibitem[{Ding et~al.(2020)Ding, Liu, Chang, Sui, and
  Gimpel}]{ding2020generative}
Xiaoan Ding, Tianyu Liu, Baobao Chang, Zhifang Sui, and Kevin Gimpel. 2020.
\newblock \href {http://arxiv.org/abs/2010.03760} {Discriminatively-tuned
  generative classifiers for robust natural language inference}.
\newblock \emph{CoRR}, abs/2010.03760.

\bibitem[{Feng et~al.(2019)Feng, Wallace, and Boyd-Graber}]{feng2019misleading}
Shi Feng, Eric Wallace, and Jordan Boyd-Graber. 2019.
\newblock \href {https://doi.org/10.18653/v1/P19-1554} {Misleading failures of
  partial-input baselines}.
\newblock In \emph{Proceedings of the 57th Annual Meeting of the Association
  for Computational Linguistics}, pages 5533--5538, Florence, Italy.
  Association for Computational Linguistics.

\bibitem[{Geiger et~al.(2018)Geiger, Cases, Karttunen, and
  Potts}]{geiger2018stress}
Atticus Geiger, Ignacio Cases, Lauri Karttunen, and Christopher Potts. 2018.
\newblock \href {http://arxiv.org/abs/1810.13033} {Stress-testing neural models
  of natural language inference with multiply-quantified sentences}.
\newblock \emph{CoRR}, abs/1810.13033.

\bibitem[{Glockner et~al.(2018)Glockner, Shwartz, and Goldberg}]{GlocknerSG18}
Max Glockner, Vered Shwartz, and Yoav Goldberg. 2018.
\newblock \href {https://doi.org/10.18653/v1/P18-2103} {Breaking {NLI} systems
  with sentences that require simple lexical inferences}.
\newblock In \emph{Proceedings of the 56th Annual Meeting of the Association
  for Computational Linguistics (Volume 2: Short Papers)}, pages 650--655,
  Melbourne, Australia. Association for Computational Linguistics.

\bibitem[{Goyal et~al.(2017)Goyal, Khot, Summers{-}Stay, Batra, and
  Parikh}]{goyal2017making}
Yash Goyal, Tejas Khot, Douglas Summers{-}Stay, Dhruv Batra, and Devi Parikh.
  2017.
\newblock \href {https://doi.org/10.1109/CVPR.2017.670} {Making the {V} in
  {VQA} matter: Elevating the role of image understanding in visual question
  answering}.
\newblock In \emph{2017 {IEEE} Conference on Computer Vision and Pattern
  Recognition, {CVPR} 2017, Honolulu, HI, USA, July 21-26, 2017}, pages
  6325--6334. {IEEE} Computer Society.

\bibitem[{Gururangan et~al.(2018)Gururangan, Swayamdipta, Levy, Schwartz,
  Bowman, and Smith}]{gururangan2018annotation}
Suchin Gururangan, Swabha Swayamdipta, Omer Levy, Roy Schwartz, Samuel Bowman,
  and Noah~A. Smith. 2018.
\newblock \href {https://doi.org/10.18653/v1/N18-2017} {Annotation artifacts in
  natural language inference data}.
\newblock In \emph{Proceedings of the 2018 Conference of the North {A}merican
  Chapter of the Association for Computational Linguistics: Human Language
  Technologies, Volume 2 (Short Papers)}, pages 107--112, New Orleans,
  Louisiana. Association for Computational Linguistics.

\bibitem[{He et~al.(2019)He, Zha, and Wang}]{he2019unlearn}
He~He, Sheng Zha, and Haohan Wang. 2019.
\newblock \href {https://doi.org/10.18653/v1/D19-6115} {Unlearn dataset bias in
  natural language inference by fitting the residual}.
\newblock In \emph{Proceedings of the 2nd Workshop on Deep Learning Approaches
  for Low-Resource NLP (DeepLo 2019)}, pages 132--142, Hong Kong, China.
  Association for Computational Linguistics.

\bibitem[{Hu et~al.(2019)Hu, Khayrallah, Culkin, Xia, Chen, Post, and
  Van~Durme}]{hu2019improved}
J.~Edward Hu, Huda Khayrallah, Ryan Culkin, Patrick Xia, Tongfei Chen, Matt
  Post, and Benjamin Van~Durme. 2019.
\newblock \href {https://doi.org/10.18653/v1/N19-1090} {Improved lexically
  constrained decoding for translation and monolingual rewriting}.
\newblock In \emph{Proceedings of the 2019 Conference of the North {A}merican
  Chapter of the Association for Computational Linguistics: Human Language
  Technologies, Volume 1 (Long and Short Papers)}, pages 839--850, Minneapolis,
  Minnesota. Association for Computational Linguistics.

\bibitem[{Iyyer et~al.(2018)Iyyer, Wieting, Gimpel, and
  Zettlemoyer}]{iyyer2018adversarial}
Mohit Iyyer, John Wieting, Kevin Gimpel, and Luke Zettlemoyer. 2018.
\newblock \href {https://doi.org/10.18653/v1/N18-1170} {Adversarial example
  generation with syntactically controlled paraphrase networks}.
\newblock In \emph{Proceedings of the 2018 Conference of the North {A}merican
  Chapter of the Association for Computational Linguistics: Human Language
  Technologies, Volume 1 (Long Papers)}, pages 1875--1885, New Orleans,
  Louisiana. Association for Computational Linguistics.

\bibitem[{Kaushik et~al.(2020)Kaushik, Hovy, and Lipton}]{kaushik2019learning}
Divyansh Kaushik, Eduard~H. Hovy, and Zachary~Chase Lipton. 2020.
\newblock \href {https://openreview.net/forum?id=Sklgs0NFvr} {Learning the
  difference that makes {A} difference with counterfactually-augmented data}.
\newblock In \emph{8th International Conference on Learning Representations,
  {ICLR} 2020, Addis Ababa, Ethiopia, April 26-30, 2020}. OpenReview.net.

\bibitem[{Khot et~al.(2018)Khot, Sabharwal, and Clark}]{khot2018scitail}
Tushar Khot, Ashish Sabharwal, and Peter Clark. 2018.
\newblock \href
  {https://www.aaai.org/ocs/index.php/AAAI/AAAI18/paper/view/17368} {Scitail:
  {A} textual entailment dataset from science question answering}.
\newblock In \emph{Proceedings of the Thirty-Second {AAAI} Conference on
  Artificial Intelligence, (AAAI-18), the 30th innovative Applications of
  Artificial Intelligence (IAAI-18), and the 8th {AAAI} Symposium on
  Educational Advances in Artificial Intelligence (EAAI-18), New Orleans,
  Louisiana, USA, February 2-7, 2018}, pages 5189--5197. {AAAI} Press.

\bibitem[{Levy et~al.(2015)Levy, Remus, Biemann, and
  Dagan}]{levy2015supervised}
Omer Levy, Steffen Remus, Chris Biemann, and Ido Dagan. 2015.
\newblock \href {https://doi.org/10.3115/v1/N15-1098} {Do supervised
  distributional methods really learn lexical inference relations?}
\newblock In \emph{Proceedings of the 2015 Conference of the North {A}merican
  Chapter of the Association for Computational Linguistics: Human Language
  Technologies}, pages 970--976, Denver, Colorado. Association for
  Computational Linguistics.

\bibitem[{Liu et~al.(2020)Liu, Xin, Chang, and Sui}]{liu2020hyponli}
Tianyu Liu, Zheng Xin, Baobao Chang, and Zhifang Sui. 2020.
\newblock \href {https://www.aclweb.org/anthology/2020.lrec-1.846}
  {{H}ypo{NLI}: Exploring the artificial patterns of hypothesis-only bias in
  natural language inference}.
\newblock In \emph{Proceedings of the 12th Language Resources and Evaluation
  Conference}, pages 6852--6860, Marseille, France. European Language Resources
  Association.

\bibitem[{Liu et~al.(2019)Liu, Ott, Goyal, Du, Joshi, Chen, Levy, Lewis,
  Zettlemoyer, and Stoyanov}]{liu2019roberta}
Yinhan Liu, Myle Ott, Naman Goyal, Jingfei Du, Mandar Joshi, Danqi Chen, Omer
  Levy, Mike Lewis, Luke Zettlemoyer, and Veselin Stoyanov. 2019.
\newblock \href {http://arxiv.org/abs/1907.11692} {Ro{BERT}a: {A} robustly
  optimized {BERT} pretraining approach}.
\newblock \emph{CoRR}, abs/1907.11692.

\bibitem[{Marelli et~al.(2014)Marelli, Bentivogli, Baroni, Bernardi, Menini,
  and Zamparelli}]{marelli2014semeval}
Marco Marelli, Luisa Bentivogli, Marco Baroni, Raffaella Bernardi, Stefano
  Menini, and Roberto Zamparelli. 2014.
\newblock \href {https://doi.org/10.3115/v1/S14-2001} {{S}em{E}val-2014 task 1:
  Evaluation of compositional distributional semantic models on full sentences
  through semantic relatedness and textual entailment}.
\newblock In \emph{Proceedings of the 8th International Workshop on Semantic
  Evaluation ({S}em{E}val 2014)}, pages 1--8, Dublin, Ireland. Association for
  Computational Linguistics.

\bibitem[{Matthews(1975)}]{matthews1975comparison}
Brian~W Matthews. 1975.
\newblock Comparison of the predicted and observed secondary structure of t4
  phage lysozyme.
\newblock \emph{Biochimica et Biophysica Acta (BBA)-Protein Structure},
  405(2):442--451.

\bibitem[{McCoy et~al.(2019)McCoy, Pavlick, and Linzen}]{mccoy2019right}
Tom McCoy, Ellie Pavlick, and Tal Linzen. 2019.
\newblock \href {https://doi.org/10.18653/v1/P19-1334} {Right for the wrong
  reasons: Diagnosing syntactic heuristics in natural language inference}.
\newblock In \emph{Proceedings of the 57th Annual Meeting of the Association
  for Computational Linguistics}, pages 3428--3448, Florence, Italy.
  Association for Computational Linguistics.

\bibitem[{Minervini and Riedel(2018)}]{minervini2018adversarially}
Pasquale Minervini and Sebastian Riedel. 2018.
\newblock \href {https://doi.org/10.18653/v1/K18-1007} {Adversarially
  regularising neural {NLI} models to integrate logical background knowledge}.
\newblock In \emph{Proceedings of the 22nd Conference on Computational Natural
  Language Learning}, pages 65--74, Brussels, Belgium. Association for
  Computational Linguistics.

\bibitem[{Naik et~al.(2018)Naik, Ravichander, Sadeh, Rose, and
  Neubig}]{DBLP:conf/coling/NaikRSRN18}
Aakanksha Naik, Abhilasha Ravichander, Norman Sadeh, Carolyn Rose, and Graham
  Neubig. 2018.
\newblock \href {https://www.aclweb.org/anthology/C18-1198} {Stress test
  evaluation for natural language inference}.
\newblock In \emph{Proceedings of the 27th International Conference on
  Computational Linguistics}, pages 2340--2353, Santa Fe, New Mexico, USA.
  Association for Computational Linguistics.

\bibitem[{Ng et~al.(2019)Ng, Yee, Baevski, Ott, Auli, and
  Edunov}]{ng2019facebook}
Nathan Ng, Kyra Yee, Alexei Baevski, Myle Ott, Michael Auli, and Sergey Edunov.
  2019.
\newblock \href {https://doi.org/10.18653/v1/W19-5333} {{F}acebook {FAIR}{'}s
  {WMT}19 news translation task submission}.
\newblock In \emph{Proceedings of the Fourth Conference on Machine Translation
  (Volume 2: Shared Task Papers, Day 1)}, pages 314--319, Florence, Italy.
  Association for Computational Linguistics.

\bibitem[{Nie et~al.(2019)Nie, Wang, and Bansal}]{nie2018analyzing}
Yixin Nie, Yicheng Wang, and Mohit Bansal. 2019.
\newblock \href {https://doi.org/10.1609/aaai.v33i01.33016867} {Analyzing
  compositionality-sensitivity of {NLI} models}.
\newblock In \emph{The Thirty-Third {AAAI} Conference on Artificial
  Intelligence, {AAAI} 2019, The Thirty-First Innovative Applications of
  Artificial Intelligence Conference, {IAAI} 2019, The Ninth {AAAI} Symposium
  on Educational Advances in Artificial Intelligence, {EAAI} 2019, Honolulu,
  Hawaii, USA, January 27 - February 1, 2019}, pages 6867--6874. {AAAI} Press.

\bibitem[{Nie et~al.(2020)Nie, Williams, Dinan, Bansal, Weston, and
  Kiela}]{nie2019adversarial}
Yixin Nie, Adina Williams, Emily Dinan, Mohit Bansal, Jason Weston, and Douwe
  Kiela. 2020.
\newblock \href {https://doi.org/10.18653/v1/2020.acl-main.441} {Adversarial
  {NLI}: A new benchmark for natural language understanding}.
\newblock In \emph{Proceedings of the 58th Annual Meeting of the Association
  for Computational Linguistics}, pages 4885--4901, Online. Association for
  Computational Linguistics.

\bibitem[{Parikh et~al.(2016)Parikh, T{\"a}ckstr{\"o}m, Das, and
  Uszkoreit}]{parikh2016decomposable}
Ankur Parikh, Oscar T{\"a}ckstr{\"o}m, Dipanjan Das, and Jakob Uszkoreit. 2016.
\newblock \href {https://doi.org/10.18653/v1/D16-1244} {A decomposable
  attention model for natural language inference}.
\newblock In \emph{Proceedings of the 2016 Conference on Empirical Methods in
  Natural Language Processing}, pages 2249--2255, Austin, Texas. Association
  for Computational Linguistics.

\bibitem[{Pennington et~al.(2014)Pennington, Socher, and
  Manning}]{pennington2014glove}
Jeffrey Pennington, Richard Socher, and Christopher Manning. 2014.
\newblock \href {https://doi.org/10.3115/v1/D14-1162} {{G}lo{V}e: Global
  vectors for word representation}.
\newblock In \emph{Proceedings of the 2014 Conference on Empirical Methods in
  Natural Language Processing ({EMNLP})}, pages 1532--1543, Doha, Qatar.
  Association for Computational Linguistics.

\bibitem[{Peters et~al.(2018)Peters, Neumann, Iyyer, Gardner, Clark, Lee, and
  Zettlemoyer}]{peters2018deep}
Matthew Peters, Mark Neumann, Mohit Iyyer, Matt Gardner, Christopher Clark,
  Kenton Lee, and Luke Zettlemoyer. 2018.
\newblock \href {https://doi.org/10.18653/v1/N18-1202} {Deep contextualized
  word representations}.
\newblock In \emph{Proceedings of the 2018 Conference of the North {A}merican
  Chapter of the Association for Computational Linguistics: Human Language
  Technologies, Volume 1 (Long Papers)}, pages 2227--2237, New Orleans,
  Louisiana. Association for Computational Linguistics.

\bibitem[{Poliak et~al.(2018{\natexlab{a}})Poliak, Haldar, Rudinger, Hu,
  Pavlick, White, and Van~Durme}]{poliak2018collecting}
Adam Poliak, Aparajita Haldar, Rachel Rudinger, J.~Edward Hu, Ellie Pavlick,
  Aaron~Steven White, and Benjamin Van~Durme. 2018{\natexlab{a}}.
\newblock \href {https://doi.org/10.18653/v1/D18-1007} {Collecting diverse
  natural language inference problems for sentence representation evaluation}.
\newblock In \emph{Proceedings of the 2018 Conference on Empirical Methods in
  Natural Language Processing}, pages 67--81, Brussels, Belgium. Association
  for Computational Linguistics.

\bibitem[{Poliak et~al.(2018{\natexlab{b}})Poliak, Naradowsky, Haldar,
  Rudinger, and Van~Durme}]{poliak2018hypothesis}
Adam Poliak, Jason Naradowsky, Aparajita Haldar, Rachel Rudinger, and Benjamin
  Van~Durme. 2018{\natexlab{b}}.
\newblock \href {https://doi.org/10.18653/v1/S18-2023} {Hypothesis only
  baselines in natural language inference}.
\newblock In \emph{Proceedings of the Seventh Joint Conference on Lexical and
  Computational Semantics}, pages 180--191, New Orleans, Louisiana. Association
  for Computational Linguistics.

\bibitem[{Ravichander et~al.(2019)Ravichander, Naik, Rose, and
  Hovy}]{ravichander2019equate}
Abhilasha Ravichander, Aakanksha Naik, Carolyn Rose, and Eduard Hovy. 2019.
\newblock \href {https://doi.org/10.18653/v1/K19-1033} {{EQUATE}: A benchmark
  evaluation framework for quantitative reasoning in natural language
  inference}.
\newblock In \emph{Proceedings of the 23rd Conference on Computational Natural
  Language Learning (CoNLL)}, pages 349--361, Hong Kong, China. Association for
  Computational Linguistics.

\bibitem[{Romanov and Shivade(2018)}]{romanov2018lessons}
Alexey Romanov and Chaitanya Shivade. 2018.
\newblock \href {https://doi.org/10.18653/v1/D18-1187} {Lessons from natural
  language inference in the clinical domain}.
\newblock In \emph{Proceedings of the 2018 Conference on Empirical Methods in
  Natural Language Processing}, pages 1586--1596, Brussels, Belgium.
  Association for Computational Linguistics.

\bibitem[{Rudinger et~al.(2017)Rudinger, May, and Van~Durme}]{RudingerMD17}
Rachel Rudinger, Chandler May, and Benjamin Van~Durme. 2017.
\newblock \href {https://doi.org/10.18653/v1/W17-1609} {Social bias in elicited
  natural language inferences}.
\newblock In \emph{Proceedings of the First {ACL} Workshop on Ethics in Natural
  Language Processing}, pages 74--79, Valencia, Spain. Association for
  Computational Linguistics.

\bibitem[{Sanchez et~al.(2018)Sanchez, Mitchell, and
  Riedel}]{sanchez2018behavior}
Ivan Sanchez, Jeff Mitchell, and Sebastian Riedel. 2018.
\newblock \href {https://doi.org/10.18653/v1/N18-1179} {Behavior analysis of
  {NLI} models: Uncovering the influence of three factors on robustness}.
\newblock In \emph{Proceedings of the 2018 Conference of the North {A}merican
  Chapter of the Association for Computational Linguistics: Human Language
  Technologies, Volume 1 (Long Papers)}, pages 1975--1985, New Orleans,
  Louisiana. Association for Computational Linguistics.

\bibitem[{Schluter and Varab(2018)}]{schluter2018data}
Natalie Schluter and Daniel Varab. 2018.
\newblock \href {https://doi.org/10.18653/v1/D18-1534} {When data permutations
  are pathological: the case of neural natural language inference}.
\newblock In \emph{Proceedings of the 2018 Conference on Empirical Methods in
  Natural Language Processing}, pages 4935--4939, Brussels, Belgium.
  Association for Computational Linguistics.

\bibitem[{Schwartz et~al.(2017)Schwartz, Sap, Konstas, Zilles, Choi, and
  Smith}]{DBLP:conf/conll/SchwartzSKZCS17}
Roy Schwartz, Maarten Sap, Ioannis Konstas, Leila Zilles, Yejin Choi, and
  Noah~A. Smith. 2017.
\newblock \href {https://doi.org/10.18653/v1/K17-1004} {The effect of different
  writing tasks on linguistic style: A case study of the {ROC} story cloze
  task}.
\newblock In \emph{Proceedings of the 21st Conference on Computational Natural
  Language Learning ({C}o{NLL} 2017)}, pages 15--25, Vancouver, Canada.
  Association for Computational Linguistics.

\bibitem[{Tsuchiya(2018)}]{tsuchiya2018performance}
Masatoshi Tsuchiya. 2018.
\newblock \href {https://www.aclweb.org/anthology/L18-1239} {Performance impact
  caused by hidden bias of training data for recognizing textual entailment}.
\newblock In \emph{Proceedings of the Eleventh International Conference on
  Language Resources and Evaluation ({LREC} 2018)}, Miyazaki, Japan. European
  Language Resources Association (ELRA).

\bibitem[{Utama et~al.(2020{\natexlab{a}})Utama, Moosavi, and
  Gurevych}]{utama-etal-2020-mind}
Prasetya~Ajie Utama, Nafise~Sadat Moosavi, and Iryna Gurevych.
  2020{\natexlab{a}}.
\newblock \href {https://doi.org/10.18653/v1/2020.acl-main.770} {Mind the
  trade-off: Debiasing {NLU} models without degrading the in-distribution
  performance}.
\newblock In \emph{Proceedings of the 58th Annual Meeting of the Association
  for Computational Linguistics}, pages 8717--8729, Online. Association for
  Computational Linguistics.

\bibitem[{Utama et~al.(2020{\natexlab{b}})Utama, Moosavi, and
  Gurevych}]{DBLP:journals/corr/abs-2009-12303}
Prasetya~Ajie Utama, Nafise~Sadat Moosavi, and Iryna Gurevych.
  2020{\natexlab{b}}.
\newblock \href {http://arxiv.org/abs/2009.12303} {Towards debiasing {NLU}
  models from unknown biases}.
\newblock \emph{CoRR}, abs/2009.12303.

\bibitem[{Wang et~al.(2019{\natexlab{a}})Wang, Pruksachatkun, Nangia, Singh,
  Michael, Hill, Levy, and Bowman}]{wang2019superglue}
Alex Wang, Yada Pruksachatkun, Nikita Nangia, Amanpreet Singh, Julian Michael,
  Felix Hill, Omer Levy, and Samuel~R. Bowman. 2019{\natexlab{a}}.
\newblock \href
  {http://papers.nips.cc/paper/8589-superglue-a-stickier-benchmark-for-general-purpose-language-understanding-systems}
  {Superglue: {A} stickier benchmark for general-purpose language understanding
  systems}.
\newblock In \emph{Advances in Neural Information Processing Systems 32: Annual
  Conference on Neural Information Processing Systems 2019, NeurIPS 2019, 8-14
  December 2019, Vancouver, BC, Canada}, pages 3261--3275.

\bibitem[{Wang et~al.(2019{\natexlab{b}})Wang, Singh, Michael, Hill, Levy, and
  Bowman}]{wang2018glue}
Alex Wang, Amanpreet Singh, Julian Michael, Felix Hill, Omer Levy, and
  Samuel~R. Bowman. 2019{\natexlab{b}}.
\newblock \href {https://openreview.net/forum?id=rJ4km2R5t7} {{GLUE:} {A}
  multi-task benchmark and analysis platform for natural language
  understanding}.
\newblock In \emph{7th International Conference on Learning Representations,
  {ICLR} 2019, New Orleans, LA, USA, May 6-9, 2019}. OpenReview.net.

\bibitem[{Wang et~al.(2019{\natexlab{c}})Wang, Sun, and Xing}]{wang2019if}
Haohan Wang, Da~Sun, and Eric~P. Xing. 2019{\natexlab{c}}.
\newblock \href {https://doi.org/10.1609/aaai.v33i01.33017136} {What if we
  simply swap the two text fragments? {A} straightforward yet effective way to
  test the robustness of methods to confounding signals in nature language
  inference tasks}.
\newblock In \emph{The Thirty-Third {AAAI} Conference on Artificial
  Intelligence, {AAAI} 2019, The Thirty-First Innovative Applications of
  Artificial Intelligence Conference, {IAAI} 2019, The Ninth {AAAI} Symposium
  on Educational Advances in Artificial Intelligence, {EAAI} 2019, Honolulu,
  Hawaii, USA, January 27 - February 1, 2019}, pages 7136--7143. {AAAI} Press.

\bibitem[{Wieting and Gimpel(2018)}]{wieting2017paranmt}
John Wieting and Kevin Gimpel. 2018.
\newblock \href {https://doi.org/10.18653/v1/P18-1042} {{P}ara{NMT}-50{M}:
  Pushing the limits of paraphrastic sentence embeddings with millions of
  machine translations}.
\newblock In \emph{Proceedings of the 56th Annual Meeting of the Association
  for Computational Linguistics (Volume 1: Long Papers)}, pages 451--462,
  Melbourne, Australia. Association for Computational Linguistics.

\bibitem[{Williams et~al.(2018)Williams, Nangia, and Bowman}]{mnli:N18-1101}
Adina Williams, Nikita Nangia, and Samuel Bowman. 2018.
\newblock \href {https://doi.org/10.18653/v1/N18-1101} {A broad-coverage
  challenge corpus for sentence understanding through inference}.
\newblock In \emph{Proceedings of the 2018 Conference of the North {A}merican
  Chapter of the Association for Computational Linguistics: Human Language
  Technologies, Volume 1 (Long Papers)}, pages 1112--1122, New Orleans,
  Louisiana. Association for Computational Linguistics.

\bibitem[{Wolf et~al.(2019)Wolf, Debut, Sanh, Chaumond, Delangue, Moi, Cistac,
  Rault, Louf, Funtowicz, and Brew}]{wolf1910huggingface}
Thomas Wolf, Lysandre Debut, Victor Sanh, Julien Chaumond, Clement Delangue,
  Anthony Moi, Pierric Cistac, Tim Rault, R{\'{e}}mi Louf, Morgan Funtowicz,
  and Jamie Brew. 2019.
\newblock \href {http://arxiv.org/abs/1910.03771} {Huggingface's transformers:
  State-of-the-art natural language processing}.
\newblock \emph{CoRR}, abs/1910.03771.

\bibitem[{Wu et~al.(2020)Wu, Moosavi, R{\"{u}}ckl{\'{e}}, and
  Gurevych}]{DBLP:journals/corr/abs-2010-03338}
Mingzhu Wu, Nafise~Sadat Moosavi, Andreas R{\"{u}}ckl{\'{e}}, and Iryna
  Gurevych. 2020.
\newblock \href {http://arxiv.org/abs/2010.03338} {Improving {QA}
  generalization by concurrent modeling of multiple biases}.
\newblock \emph{CoRR}, abs/2010.03338.

\bibitem[{Yaghoobzadeh et~al.(2019)Yaghoobzadeh, des Combes, Hazen, and
  Sordoni}]{yaghoobzadeh2019robust}
Yadollah Yaghoobzadeh, Remi~Tachet des Combes, Timothy~J. Hazen, and Alessandro
  Sordoni. 2019.
\newblock \href {http://arxiv.org/abs/1911.03861} {Robust natural language
  inference models with example forgetting}.
\newblock \emph{CoRR}, abs/1911.03861.

\bibitem[{Yang et~al.(2019)Yang, Dai, Yang, Carbonell, Salakhutdinov, and
  Le}]{yang2019xlnet}
Zhilin Yang, Zihang Dai, Yiming Yang, Jaime~G. Carbonell, Ruslan Salakhutdinov,
  and Quoc~V. Le. 2019.
\newblock \href
  {http://papers.nips.cc/paper/8812-xlnet-generalized-autoregressive-pretraining-for-language-understanding}
  {{XLN}et: Generalized autoregressive pretraining for language understanding}.
\newblock In \emph{Advances in Neural Information Processing Systems 32: Annual
  Conference on Neural Information Processing Systems 2019, NeurIPS 2019, 8-14
  December 2019, Vancouver, BC, Canada}, pages 5754--5764.

\bibitem[{Zhang et~al.(2019)Zhang, Bai, Liang, Bai, Chang, Yu, Zhu, and
  Zhao}]{zhang2019selection}
Guanhua Zhang, Bing Bai, Jian Liang, Kun Bai, Shiyu Chang, Mo~Yu, Conghui Zhu,
  and Tiejun Zhao. 2019.
\newblock \href {https://doi.org/10.18653/v1/P19-1435} {Selection bias
  explorations and debias methods for natural language sentence matching
  datasets}.
\newblock In \emph{Proceedings of the 57th Annual Meeting of the Association
  for Computational Linguistics}, pages 4418--4429, Florence, Italy.
  Association for Computational Linguistics.

\end{thebibliography}
\bibliographystyle{acl_natbib}

\end{document}